\documentclass[journal]{IEEEtran}

\usepackage{times}
\usepackage{soul}
\usepackage{url}
\usepackage[hidelinks]{hyperref}
\usepackage[utf8]{inputenc}
\usepackage[small]{caption}
\usepackage{graphicx}
\usepackage{amsmath}
\usepackage{amsthm}
\usepackage{booktabs}
\usepackage{algorithm}
\usepackage{algorithmic}
\usepackage{multicol}
\usepackage{subfigure}
\usepackage{float}
\usepackage{standalone}
\usepackage{amsfonts}
\usepackage{enumitem}
\usepackage{multirow}
\usepackage[normalem]{ulem}

\author{Yudi~Dong,~\IEEEmembership{Graduate Student Member,~IEEE,}
        Huaxia~Wang,~\IEEEmembership{Member,~IEEE,}
        and~Yu-Dong~Yao,~\IEEEmembership{Fellow,~IEEE}\vspace{-2em}
\thanks{Y. Dong and Y.-D. Yao are with the Department of Electrical $\&$ Computer Engineering, Stevens Institute of Technology, Hoboken, New Jersey 07030, USA (e-mail: ydong6@stevens.edu;yyao@stevens.edu).

H. Wang is with the College of Engineering, Architecture and Technology (CEAT), Oklahoma State University, Stillwater, OK 74078 USA (e-mail: huaxia.wang@okstate.edu).}}

\title{\huge{A Robust Adversarial Network-Based End-to-End Communications System With Strong Generalization Ability Against Adversarial Attacks}}

\begin{document}

\maketitle

\begin{abstract}
We propose a novel defensive mechanism based on a generative adversarial network (GAN) framework to defend against adversarial attacks in end-to-end communications systems\footnote{Source Code: https://github.com/YudiDong/GAN-based-E2E-communications-system-for-defense-against-adversarial-attack}.
Specifically, we utilize a generative network to model a powerful adversary and enable the end-to-end communications system to combat the generative attack network via a minimax game.
We show that the proposed system not only works well against white-box and black-box adversarial attacks but also possesses excellent generalization capabilities to maintain good performance under no attacks.
We also show that our GAN-based end-to-end system outperforms the conventional communications system and the end-to-end communications system with/without adversarial training.
\end{abstract}


\section{Introduction}
\label{sec:intro}

Deep neural networks (DNNs) bring wireless communications into a new era of deep learning and artificial intelligence.
One of the insightful ideas is end-to-end learning of communications systems~\cite{o2017introduction}, which re-designs the physical layer by employing a neural network instead of multiple independent blocks at the transmitter and the receiver.
Particularly, an autoencoder architecture~\cite{goodfellow2016deep} is utilized for end-to-end communications, where an encoder neural network (NN) and a decoder NN are respectively utilized in the transmitter and receiver to replace signal processing tasks. 
Through jointly training the transmitter NN and the receiver NN, the end-to-end communications system can achieve global optimization and considerable performance improvements~\cite{o2017introduction}.


However, neural networks have an inherent/natural vulnerability to adversarial attacks~\cite{goodfellow2014explaining}. 
That is, a neural network model can easily lead to a false output by adding a small perturbation into the input of a neural network.
Such perturbation, called adversarial perturbation, is an elaborate vector designed based on the receptive fields of inputs in the neural network model.
This vulnerability threatens almost all deep learning-based systems including the end-to-end learning based communications system in terms of robustness and security. 
A recent work~\cite{sadeghi2019physical} investigates adversarial attacks against
autoencoder end-to-end communications systems, which crafts universal adversarial perturbations using a fast gradient method (FGM)~\cite{goodfellow2014explaining}. By leveraging the broadcast nature of the wireless channel, attackers can inject adversarial perturbations into the input of the receiver NN, which causes a more significantly negative impact on the end-to-end learning based systems than conventional communications systems~\cite{sadeghi2019physical}.

A direct defensive method against adversarial attacks is to train the end-to-end system with adversarial perturbations, which is called adversarial training~\cite{goodfellow2014explaining}.  
However, adversarial training only works for some specific adversarial perturbations that have been added to the training. For other various and new adversarial perturbations, adversarial training may be incapable of any defense~\cite{tramer2019adversarial}. Also, adversarial training degrades the generalization ability of neural networks~\cite{raghunathan2019adversarial}, which can lead to the poor performance of neural networks on unperturbed/clean inputs. 
Therefore, a more effective defense mechanism is desired for robust deep learning of end-to-end communications systems. 

To this end, in this paper, we propose to integrate the GAN framework~\cite{NIPS2014gan} into the autoencoder based end-to-end communications system for defense against various adversarial attacks.
We utilize a generative network as an adversary to generate adversarial perturbations that can fool the receiver NN into recovering the false message. 
By leveraging the great computational capacity of the neural network, the generative network can generate various and powerful perturbations. 
Meanwhile, the discriminative network is the decoder NN in the end-to-end system, which is responsible for recovering the correct message from both clean signal and perturbed signal with adversarial perturbations generated by the generative network. 
The generative network and the discriminative network are trained in a confrontation game, where the generative network becomes a powerful adversary while the discriminative network (i.e., decoder NN) becomes a robust defender.

The main contributions of our paper are as follows.
\begin{itemize}
\item This work is the first to resolve the security and robustness issue induced by adversarial attacks in the end-to-end communications system, where we build a robust and defensive GAN-based end-to-end communications system by jointly and adversarially training an autoencoder network against a generative attack network.
\item Unlike the adversarial training method that is hard to gain simultaneously defense and generalization capacity, the proposed approach can effectively defend against various adversarial attacks including white-box attacks and black-box attacks and, meanwhile, it has excellent generalization performance to remain in low error rates on clean inputs. 
\item Consensus optimization is utilized in the training of the proposed end-to-end system, which ensures a stable and impartial minimax game to train a defensive end-to-end communications system.
\end{itemize}

\section{Preliminaries}
\label{sec:pre}

In this section, we introduce the preliminary studies regarding autoencoder based end-to-end communications systems and adversarial attacks. Also, we discuss the attack model and the method of crafting adversarial perturbations for attacking an end-to-end communications system. 

\begin{figure}[t]
\centering
\includegraphics[width=\linewidth]{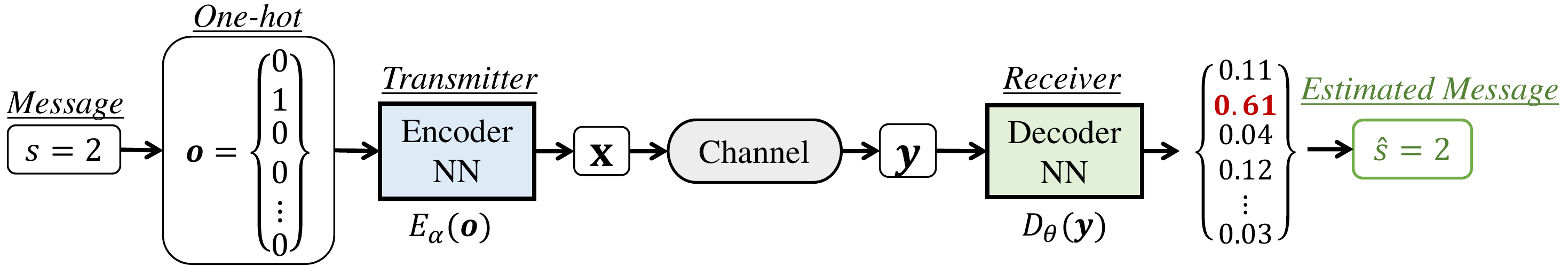}
\caption{Illustration of an end-to-end autoencoder communications system.}
\label{fig:autoencoder}
\end{figure}

\subsection{Autoencoder Based End-to-End Communications System}

%

Fig.~\ref{fig:autoencoder} illustrates a typical end-to-end autoencoder communications system~\cite{o2017introduction}, which is implemented in this paper. 
Specifically, message $s$ that needs to be transmitted is chosen from a message set $\mathcal{M}=\left \{ 1, 2, \cdots, M \right \}$, where $M=2^k$ and $k$ is the bit number of a message. 
The message $s$ is first preprocessed as a one-hot binary vector $\mathbf{o}\in\mathbb{R}^{M}$ where the $s^{th}$ element of $\mathbf{o}$ is equal to one and all others are zero.
Then the one-hot message goes through the encoder NN to perform a mapping: $E_{\alpha}: \mathcal{O} \mapsto \mathbb{R}^{2n}$, which generates the output signal $\mathbf{x} = E_{\alpha}(\mathbf{o}) \in \mathbb{R}^{2n}$, where $E_{\alpha}$ refers the encoder model parameterized by ${\alpha}$, $\mathcal{O}$ is the message set via one-hot calculation, $n$ refers to the number of channel uses and $\mathbf{x}$ is a concatenation of the real and imaginary parts of the transmitted signal. Consider the hardware constraints of a transmitter, we  restrict the energy of the transmitted signal as $\left \|\mathbf{x}  \right \|_{2}^{2}\leq \frac{n}{2}$. 
Next, an additive white Gaussian noise (AWGN) channel is used for the transmission of $\mathbf{x}$ to obtain the received signal $\mathbf{y}\in \mathbb{R}^{2n}$, where $\mathbf{y}$ involves $\mathbf{x}$ and noise. 
We assign the fixed variance $\sigma^2= \left ( 2RE_{b}/N_{0} \right)^{-1}$ in the AWGN channel, where $R = k/n$, computed by bit number $k$ and $n$ channel uses, is the data rate in our communications system. $E_{b}/N_{0}$ is the energy per bit to noise power spectral density ratio.
Finally, the decoder NN performs a mapping $D_{\theta}: \mathbb{R}^{2n} \mapsto \mathcal{M}$ to recover the estimated message $\hat{s} = D_{\theta}(\mathbf{y})$, where $D_{\theta}$ is the decoder model parameterized by ${\theta}$.
In particular, the softmax layer of the decoder NN generates the vector $(0,1)^{M}$.
The estimated message $\hat{s}$ is set as the index of the highest value in the output vector $(0,1)^{M}$.

\subsection{Adversarial Attacks}

\begin{figure}[ht]
\centering
\includegraphics[width=\linewidth]{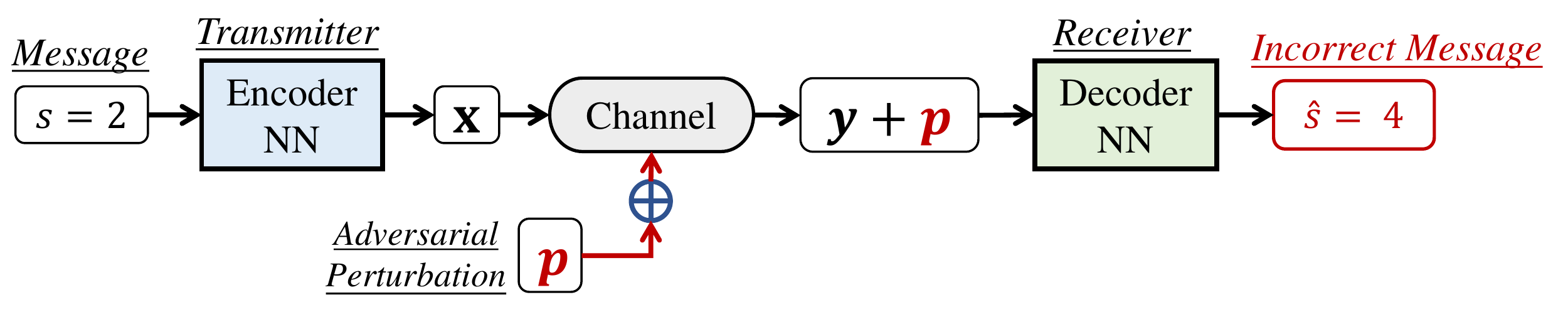}
\caption{Adversarial attacks against an end-to-end autoencoder communications system.}
\label{fig:attack}
\end{figure}

Neural networks have a natural vulnerability to adversarial attacks, where input with adversarial perturbations can lead a well-trained neural network to output a wrong answer with high confidence~\cite{goodfellow2014explaining}. 
An adversarial perturbation is a carefully crafted vector or matrix with small values, which are imperceptible but sensitive to neural networks. 
Due to this property of neural networks, the security and robustness of deep learning-based systems are compromised by adversarial attacks. 
In our case, an autoencoder based end-to-end communications system can be easily fooled by using physical adversarial attacks~\cite{sadeghi2019physical}. 
As shown in Fig.~\ref{fig:attack}, attackers can leverage the broadcast nature of the channel and emit an interference signal of adversarial perturbation $\mathbf{p}$ to the channel. The perturbed received signal $\mathbf{y}+\mathbf{p}$ forces the decoder NN to provide an incorrect output. 
Under adversarial attacks, autoencoder communications systems have more significant performance degradation than conventional communications systems~\cite{sadeghi2019physical}.
According to the knowledge of attackers, adversarial attacks can be divided into white-box attacks and black-box attacks~\cite{yuan2019adversarial}. In white-box attacks, an attacker has complete knowledge of the NN model $D_{\theta}$.
In black-box attacks, attackers only know the output of the decoder model but have no information about the NN model. 

\subsection{Attack Model: Crafting Adversarial Perturbation}

\begin{figure}[t]
\centering
\includegraphics[width=0.7\linewidth]{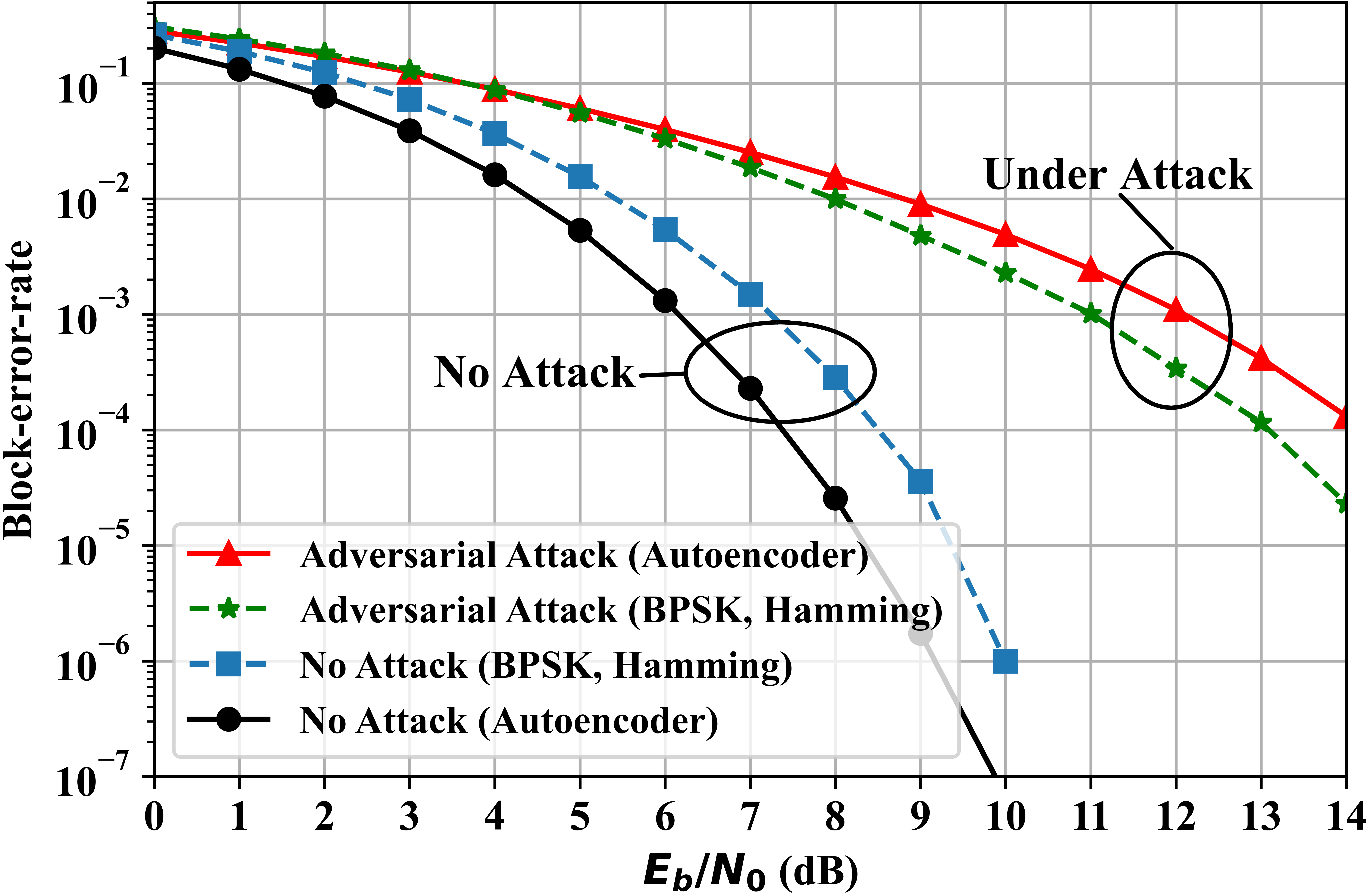}
\caption{BLER performance comparison of the autoencoder end-to-end system and conventional scheme (BPSK modulation with Hamming coding) under adversarial attacks.}
\label{fig:white_attack_ae-bpsk}
\end{figure}

\begin{figure*}[t]
\centering
\includegraphics[width=0.7\linewidth]{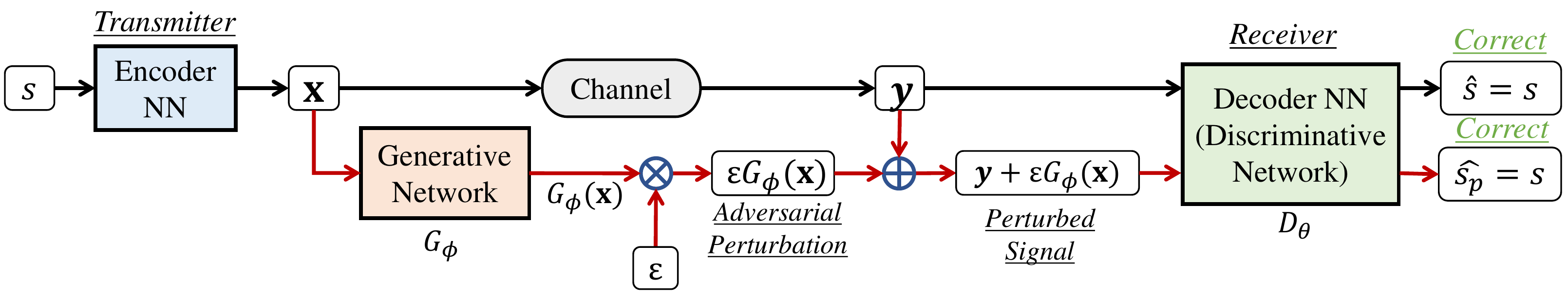}
\caption{The proposed adversarial network based approach for robust end-to-end communications system.}
\label{fig:system}
\end{figure*}

To perform white-box attacks on the decoder NN model $D_{\theta}$ that generates the estimated message $\hat{s} = D_{\theta}(\mathbf{y})$, we need to find an adversarial perturbation $\mathbf{p}$ such that $\mathbf{y}+\mathbf{p}$ results in an incorrect output, which is described as
\begin{equation}
\label{eq:attack_model}
\underset{\mathbf{p}}{\arg \min\left \| \mathbf{p} \right \|_{2}}\quad\text{s.t.}\ \arg \max D_{\theta}(\mathbf{y}+\mathbf{p})\neq D_{\theta}(\mathbf{y}).
\end{equation}
To solve the problem (\ref{eq:attack_model}) of generating adversarial perturbations, the FGM method~\cite{goodfellow2014explaining} is commonly used to obtain an optimal $\ell_{2}$-norm constrained perturbation, 
\begin{equation}
\label{eq:fgs}
\mathbf{p} = \epsilon \cdot \frac{\nabla_{\mathbf{y}}l(D_{\theta}(\mathbf{y}),s)}{\left \| \nabla_{\mathbf{y}}l(D_{\theta}(\mathbf{y}),s) \right \|_{2}},
\end{equation}
where $\epsilon$ is a small scaling coefficient, $l$ denotes the loss function, and $\nabla_{\mathbf{y}}$ is the gradient of the loss function $l$ with respect to the input $\mathbf{y}$. 
However, FGM requires the knowledge of the message $s$ that is unknown to the transmission process.
Therefore, Sadeghi~\textit{et al.} introduce an input-agnostic FGM~\cite{sadeghi2019physical} to generate an universal perturbation $\mathbf{p}$ that works for all messages from $\mathcal{M}$. 
This method is used in this paper for crafting adversarial perturbations.
For the black-box attacks, attackers cannot obtain any information about our autoencoder. Thus, attackers need to design white-box perturbations based on a substitute autoencoder system that is fully open to attackers. These adversarial perturbations are also effective for other unknown autoencoder systems due to the transferability of adversarial attacks~\cite{yuan2019adversarial}. We use this general approach to perform black-box attacks on the autoencoder system and the proposed system.


Fig.~\ref{fig:white_attack_ae-bpsk} shows the block-error-rate (BLER) of an autoencoder end-to-end communications system with $n=7$ channel uses and $k=4$ bits per channel, and the BLER of a conventional communications system using binary phase-shift keying (BPSK) modulation and Hamming (7,4) code with hard-decision (HD) decoding~\cite{o2017introduction, sadeghi2019physical}. The BLER is calculated as the ratio of $\hat{s}\neq s$. The smaller BLER indicates the better system performance. 
We can see that the autoencoder outperforms the conventional scheme if there is no attack. However, by performing the adversarial attacks using the input-agnostic FGM, the performance of the autoencoder is degraded more significantly, where the performance of the autoencoder  is worse than the conventional scheme. 
This paper is to address this issue induced by adversarial attacks in the end-to-end communications system.

\section{End-to-End Communications System Using Adversarial Networks}
\label{sec:method}


We integrate the GAN framework~\cite{NIPS2014gan, Wang2019ICLR} into an autoencoder communications system. 
As shown in Fig.~\ref{fig:system}, a neural network $G_{\phi}$ parameterized by ${\phi}$ is added as the generative network. 
The decoder NN $D_{\theta}$ is served as a discriminative network. 
In our case, the purposes of the generative and discriminative networks are different from the original GAN. 
Here the generative network acts as an adversary to model and generate adversarial perturbation (i.e., $\epsilon G_{\phi}(\mathbf{x})$) based on the input $\mathbf{x}$ and a scaling factor $\epsilon=0.2$. 
The discriminative network tries to estimate the correct message from both clean signal $\mathbf{y}$ and perturbed signal $\mathbf{y}+\epsilon G_{\phi}(\mathbf{x})$.
The generative and discriminative networks are trained jointly and adversarially, where the generative network generates evermore powerful adversarial perturbation but the discriminative network still correctly estimates messages from the heavily perturbed signals. 
With the proposed adversarial network-based approach, the autoencoder communications system obtains a strong capability to defend against adversarial attacks.

\subsection{Objective Function}
The intuition of an ideal defensive method is to find a solution $\theta$ of the decoder NN that simultaneously has the small loss $L(\theta)$ on the clean inputs and the small loss $L_{p}(\theta)$ on the inputs with adversarial perturbations,
\begin{equation}
L(\theta) = l(D_{\theta}(\mathbf{y}),s),
\end{equation}
\begin{equation}
L_{p}(\theta) = l(D_{\theta}(\mathbf{y}+p)),s).
\end{equation}
However, it is hard to find a single solution for both $L(\theta)$ and $L_{p}(\theta)$. 
There is a trade-off between $L(\theta)$ and $L_{p}(\theta)$. 
The traditional adversarial training usually satisfies either the small loss $L(\theta)$ of clean input or the small loss  $L_{p}(\theta)$ of the perturbed inputs, which causes the model $D_{\theta}$ to lose either defense ability or generalization ability. 

To satisfy the above two requirements, we try to find an optimal parameter $\theta$ of the decoder NN $D_{\theta}$ to minimize the loss between the output of the clear signal $y$ and the groundtruth $s$, as well as minimize the loss between the output of the perturbed signal $D_{\theta}(\mathbf{y}+p)$ and the groundtruth $s$, where the objective of the decoder NN is 
\begin{equation}
\arg \underset{\theta}{\min}\quad \left [l\left(D_{\theta}\left(\mathbf{y}\right),s\right) + l\left(D_{\theta}\left(\mathbf{y}+p\right),s\right)\right ].
\end{equation}
In our approach, we model the adversarial perturbation using the generative neural network $G_{\phi}$, and the objective of the decoder NN becomes
\begin{equation}
\label{eq:ds}
\arg \underset{\theta}{\min}\quad \left [l(D_{\theta}(\mathbf{y}),s) + l(D_{\theta}(\mathbf{y}+\epsilon G_{\phi}(\mathbf{x})),s)\right ],
\end{equation}
where $\epsilon G_{\phi}(\mathbf{x})$ is the generated adversarial perturbation.
In order to enable the decoder NN to handle as many perturbation types as possible, we want the generative neural network $G_{\phi}$ to be a powerful adversary, where a generative network parameter $\phi$ is trained to maximize the loss between the output of the perturbed signal $D_{\theta}(\mathbf{y}+\epsilon G_{\phi}(\mathbf{x}))$ and the groundtruth $s$, 
\begin{equation}
\label{eq:ge}
\arg \underset{\phi}{\max}\quad l(D_{\theta}(\mathbf{y}+\epsilon G_{\phi}(\mathbf{x})),s).
\end{equation}

Finally, we jointly train the decoder NN (i.e., discriminative network) and the generative network to find a solution of a minimax game between  $D_{\theta}$ and $G_{\phi}$,
\begin{equation}
\begin{aligned}
\arg \underset{\theta}{\min}\ \underset{\phi}{\max} \quad & \left [l(D_{\theta}(\mathbf{y}),s) + l(D_{\theta}(\mathbf{y}+\epsilon G_{\phi}(\mathbf{x})),s)\right ] \\ & + \left [l(D_{\theta}(\mathbf{y}+\epsilon G_{\phi}(\mathbf{x})),s) \right ],
\end{aligned}
\end{equation}
where our final objective is realized that the discriminative network is capable of countering against a powerful adversary while has a good generalization performance.

\subsection{Consensus Optimization For GAN Training}
The stability and convergence of GAN training is a very challenge task, which suffers from the problems of non-convergence, mode collapse, and diminished gradient. 
In this paper, we adopt a consensus optimization approach~\cite{mescheder2017numerics} to regularize gradients to stabilize the GAN training. 

Denote the objective of the discriminative network (i.e., Eq. (\ref{eq:ds})) as $d(\theta,\phi)$ and denote the objective of the generative network (i.e., Eq. (\ref{eq:ge})) as $g(\theta,\phi)$. The gradient vector field $v(\theta,\phi)$ of this minimax game is defined as
\begin{equation}
v(\theta,\phi)=
\begin{pmatrix}
\nabla_{\theta} d(\theta,\phi)\\ 
\nabla_{\phi} g(\theta,\phi)
\end{pmatrix}.
\end{equation}
The GAN training is to find a solution of $v(\theta,\phi)= 0$. 
However, the eigenvalues of the Jacobian of $v(\theta,\phi)$ could be zero in real part or be very large in imaginary part~\cite{mescheder2017numerics}, which results in the convergence failure of GAN training. To this end, we respectively add a regularization factor 
$\mathcal{L}(\theta,\phi) = \frac{1}{2}\left \| v(\theta,\phi) \right \|^{2}$
to the objectives of the discriminative network and the generative network. The new gradient vector field $v_{s}(\theta,\phi)$ is obtained~\cite{mescheder2017numerics}
\begin{equation}
v_s(\theta,\phi)=
\begin{pmatrix}
\nabla_{\theta} (d(\theta,\phi)-\gamma \mathcal{L}(\theta,\phi))\\ 
\nabla_{\phi} (g(\theta,\phi)-\gamma \mathcal{L}(\theta,\phi))
\end{pmatrix},
\end{equation}
where $\gamma$ is a constant parameter for regularization. This added regularization factor $\mathcal{L}(\theta,\phi)$ can help the two networks to reach a consensus optimization with better convergence.

\section{Evaluation Results}
\label{sec:eval}

In this section, we evaluate the proposed GAN based end-to-end communications system by comparing it with the conventional communications system (Section~\ref{subsec:ev_bpsk}), the autoencoder end-to-end communications system with regular training and adversarial training (Section~\ref{subsec:ev_ae}).
To examine the robustness of those systems, we calculate their BLER performance under different scenarios involving white-box attacks, black-box attacks, and no attacks. 

\subsection{Neural Network Architecture}

We implement our adversarial network based approach into two different end-to-end communications systems: a multilayer perceptron (MLP) based end-to-end communications system and a convolutional neural network (CNN) based end-to-end communications system, which are given in Table~\ref{tab:mlp_str} and Table~\ref{tab:cnn_str}, respectively. 
The encoder NN and decoder NN used in these two systems are the same as in~\cite{sadeghi2019physical}. 
For the design of the generative network, one noticed rule is that the depth (i.e., number of layers) of the generative network and the decoder NN (i.e., discriminative network) should be similar, which can reach equal competition between the generative network and the discriminative network to result in better performance.



\subsection{Experiment Setup}
In the experiments under white-box attacks, the proposed system uses the network architecture listed in Table~\ref{tab:mlp_str}.
The autoencoder system use the same MLP encoder and MLP decoder listed in Table~\ref{tab:mlp_str}.
The conventional communications system uses BPSK modulation and Hamming coding with HD decoding. 
The adversarial perturbations for attacking these three systems are generated using FGM~\cite{sadeghi2019physical} based on the MLP decoder. 
In the experiments under black-box attacks, the proposed system uses the network architecture listed in Table~\ref{tab:cnn_str}.
The autoencoder system uses the same CNN encoder and CNN decoder in Table~\ref{tab:cnn_str}.
The conventional communications system also uses BPSK modulation and Hamming coding with HD decoding. 
The adversarial perturbations for black-box attacks are generated from the MLP decoder. In addition, the proposed system and the autoencoder system are all sufficiently trained with the same hyper-parameters on TensorFlow-GPU.


\begin{table}[t]
\centering
\caption{NN Architectures used in our approach (MLP based).}
\label{tab:mlp_str}
\resizebox{0.8\linewidth}{!}{%
\begin{tabular}{|c|c|c|c|}
\hline
\multicolumn{1}{|l|}{\textbf{Name}} & \textbf{Encoder NN}                          & \textbf{Decoder NN}                  & \textbf{Generative Network} \\ \hline
\multirow{4}{*}{\textbf{Layer}} & \multirow{2}{*}{\textit{FC+eLU}} & \multirow{2}{*}{\textit{FC+ReLU}} & \textit{Conv1d+ReLU} \\ \cline{4-4} 
                                    &                                              &                                      & \textit{Cov1d+ReLU+Flatten} \\ \cline{2-4} 
                                    & \multirow{2}{*}{\textit{FC+Linear+ $\ell_{2}$ Norm}} & \multirow{2}{*}{\textit{FC+Softmax}} & \textit{FC+Linear}          \\ \cline{4-4} 
                                    &                                              &                                      & \textit{Normalization ($\ell_{2}$)} \\ \hline
\end{tabular}%
}
\end{table}

\begin{table}[t]
\centering
\caption{NN Architectures used in our approach (CNN-based).}
\label{tab:cnn_str}
\resizebox{0.9\linewidth}{!}{%
\begin{tabular}{|c|c|c|c|}
\hline
\multicolumn{1}{|l|}{\textbf{Name}} & \textbf{Encoder NN} & \textbf{Decoder NN} & \textbf{Generative Network} \\ \hline
\multirow{4}{*}{\textbf{Layer}} & \textit{FC+eLU}              & \textit{Conv2d+ReLU}         & \textit{Conv1d+ReLU+BN}        \\ \cline{2-4} 
                                & \textit{Conv1d+ReLU+Flatten} & \textit{Conv2d+ReLU+Flatten} & \textit{Cov1d+ReLU+BN+Flatten} \\ \cline{2-4} 
                                & \textit{FC+Linear}           & \textit{FC+ReLU}             & \textit{FC+Linear}             \\ \cline{2-4} 
                                & \textit{Normalization ($\ell_{2}$)}  & \textit{FC+Softmax}          & \textit{Normalization ($\ell_{2}$)}    \\ \hline
\end{tabular}%
}
\end{table}

\subsection{Proposed Approach versus Conventional Communications System}
\label{subsec:ev_bpsk}

\begin{figure}[t]
	\centering
	\subfigure[BLER under whiter-box attacks]{
		\begin{minipage}{0.7\linewidth}
			\label{fig:GAN-BPSK-wh}
			\centering
			\includegraphics[width=\linewidth]{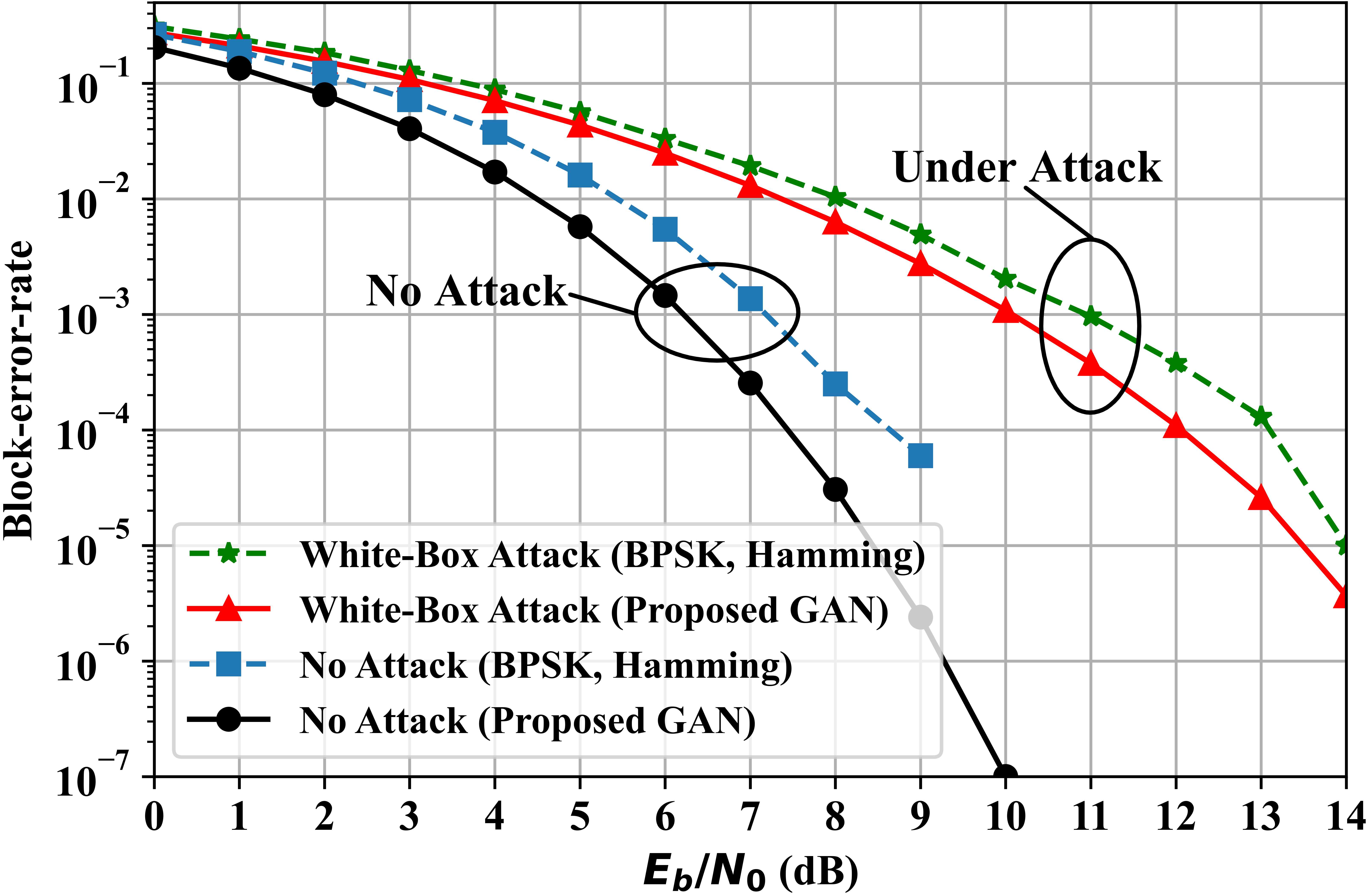}
		\end{minipage}
	}	
	\subfigure[BLER under black-box attacks]{
		\begin{minipage}{0.7\linewidth}
			\label{fig:GAN-BPSK-bl}
			\centering
			\includegraphics[width=\linewidth]{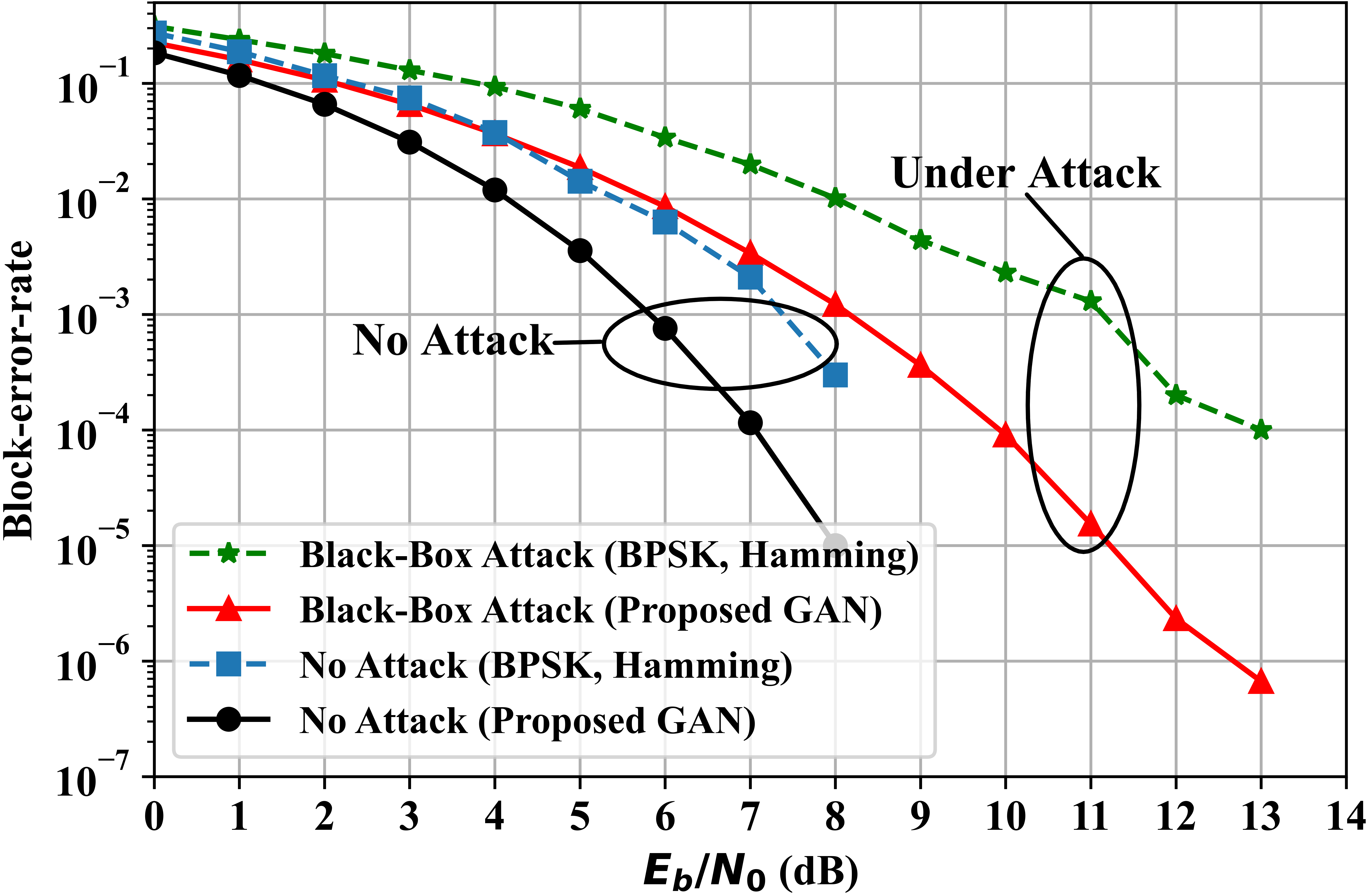}
		\end{minipage}
	}
\caption{BLER performance comparison of the proposed GAN-based end-to-end communications system and the conventional communications system.}
\label{fig:GAN-BPSK}
\end{figure}

We first compare our proposed GAN-based communications system with the conventional communications system under adversarial attacks and no attack. 
In Figure~\ref{fig:GAN-BPSK}, we can see that the performance of our proposed system is better than the conventional communications system under no attacks.
When we attack these two systems using whiter-box attacks, as shown in Figure~\ref{fig:GAN-BPSK-wh}, our system can mitigate the effect of attacks and has a better performance than the conventional communications system. 
While performing black-box attacks in Figure~\ref{fig:GAN-BPSK-bl}, our system shows a considerable defense capacity, where the performance of our system significantly outperforms the conventional one, which is very close to the performance under no attacks.

\subsection{Proposed Approach versus Autoencoder End-to-End Communications System}
\label{subsec:ev_ae}

\begin{figure}[t]
	\centering
	\subfigure[BLER under whiter-box attacks]{
		\begin{minipage}{0.7\linewidth}
			\label{fig:GAN-AE-AEAD-wh}
			\centering
			\includegraphics[width=\linewidth]{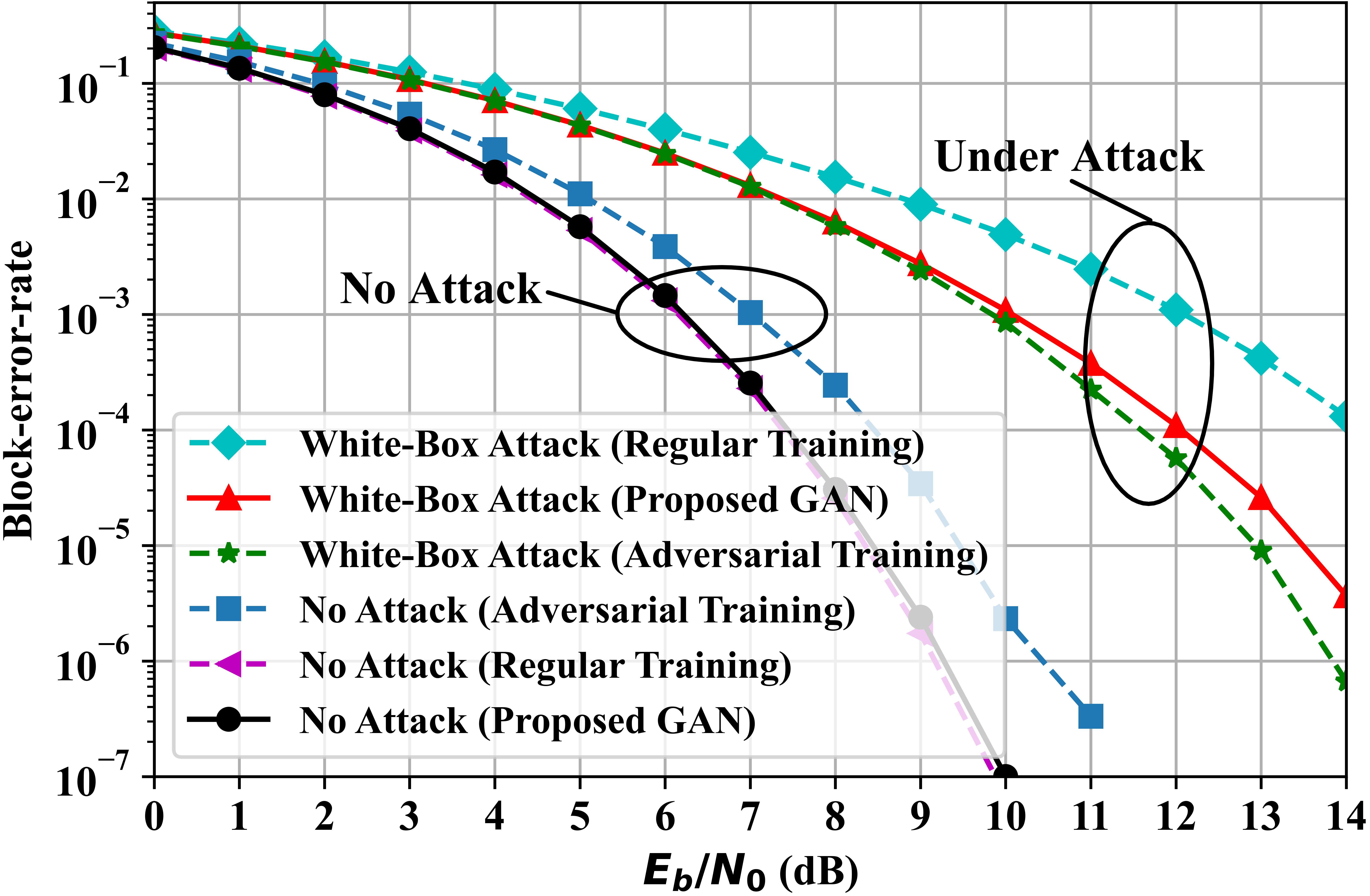}
		\end{minipage}
	}	
	\subfigure[BLER under black-box attacks]{
		\begin{minipage}{0.7\linewidth}
			\label{fig:GAN-AE-AEAD-bl}
			\centering
			\includegraphics[width=\linewidth]{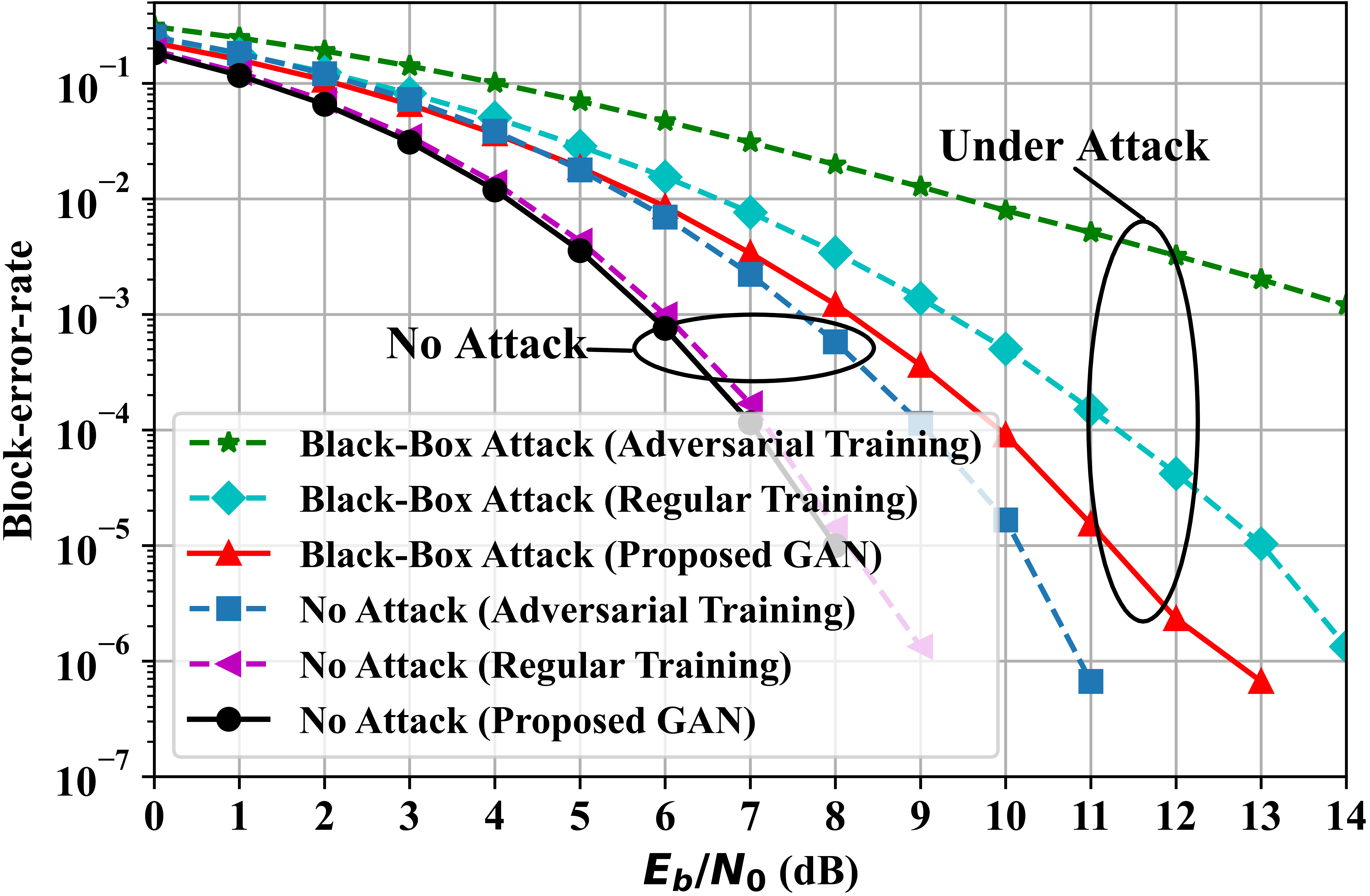}
		\end{minipage}
	}
\caption{BLER performance comparison of the proposed GAN-based end-to-end communications system and the autoencoder end-to-end communications system with regular training and adversarial training.}
\label{fig:GAN-AE-AEAD}
\end{figure}

Next, we compare our proposed system with the autoencoder end-to-end system that uses regular training and adversarial training, respectively.
Regular training means that we train the autoencoder end-to-end system using clean inputs.
Adversarial training means that we training the autoencoder system with both the clean inputs and the inputs with adversarial perturbations.
For the results of white-box attacks shown in Figure~\ref{fig:GAN-AE-AEAD-wh}, we can see that the regular training based autoencoder system has no capability to defend against white-box attacks, where the regular training has the highest error rate. 
The adversarial training based autoencoder system achieves successful defense against white-box attacks, which obtains large performance improvements compared with the regular training based autoencoder system.
The adversarial training is effective for defending against white-box perturbations because it augments the training data with the same perturbations beforehand.
Our proposed system also achieves a good performance similar to the adversarial training, indicating a good defense against white-box attacks. 
Notably, adversarial training causes considerable performance degradation when there is no attack, indicating that adversarial training degrades the generalization ability of the autoencoder. In contrast, our proposed system still remains in a good performance under no attacks with a strong generalization ability. 
For the results of black-box attacks shown in Figure~\ref{fig:GAN-AE-AEAD-bl}, our proposed system still shows a good defensive ability against black-box perturbations, but the adversarial training leads to a defense failure where the adversarial training has a high error rate. 
This is because the perturbations used for black-box attacking are different from the perturbations used in the adversarial training. Adversarial training does not work well for unknown perturbations. 
In contrast, our system can defend against various unknown perturbations.
Similarly, Figure~\ref{fig:GAN-AE-AEAD-bl} also indicates that the adversarial training shows a performance degradation under no attacks while our system shows good generalization performance.

\vspace{1.5mm}
\section{Conclusions}
\label{sec:concl}
\vspace{1mm}
This paper presents a novel GAN-based defense approach for end-to-end learning of communications systems, which uses a generative network to model powerful adversarial perturbations and jointly train the end-to-end communications system against the generative attack network.
Our approach can learn an end-to-end communication system which is robust to various adversarial perturbations including both white-box and black-box attacks, without degrading the generalization performance of the system.
In evaluation results, our GAN-based communications system shows better performance and defense capability than the classical communications scheme and the end-to-end communications system with regular training and adversarial training.


\bibliographystyle{IEEEtran}
\bibliography{ref}
\end{document}